# Probabilistic Semantic Web Mining
# Using Artificial Neural Analysis

**1. T. KRISHNA KISHORE,  2.  T.SASI VARDHAN,  AND  3. N. LAKSHMI NARAYANA**

## Abstract

Most of the web user's requirements are search or navigation time and getting correctly matched result. These constrains can be satisfied with some additional modules attached to the existing search engines and web servers. This paper proposes that powerful architecture for search engines with the title of Probabilistic Semantic Web Mining named from the methods used.

With the increase of larger and larger collection of various data resources on the World Wide Web (WWW), Web Mining has become one of the most important requirements for the web users. Web servers will store various formats of data including text, image, audio, video etc., but servers can not identify the contents of the data. These search techniques can be improved by adding some special techniques including semantic web mining and probabilistic analysis to get more accurate results.

Semantic web mining technique can provide meaningful search of data resources by eliminating useless information with mining process. In this technique web servers will maintain Meta information of each and every data resources available in that particular web server. This will help the search engine to retrieve information that is relevant to user given input string.

This paper proposing the idea of combing these two techniques Semantic web mining and Probabilistic analysis for efficient and accurate search results of web mining. SPF can be calculated by considering both semantic accuracy and syntactic accuracy of data with the input string. This will be the deciding factor for producing results.

## 1. Concept of Data Mining

### 1.1.1 Data

Data are any facts, numbers, or text that can be processed by a computer. Today, organizations are accumulating vast and growing amounts of data in different formats and different databases. This includes:

- Operational or transactional data such as, sales, cost, inventory, payroll, and accounting

- No operational data, such as industry sales, forecast data, and macro economic data

- Meta data - data about the data itself, such as logical database design or data dictionary definitions

### 1.1.2 Information

The patterns, associations, or relationships among all this *data* can provide *information*. For example, analysis of retail point of sale transaction data can yield information on which products are selling and when.

### 1.1.3 Knowledge

Information can be converted into *knowledge* about historical patterns and future trends. For example, summary information on retail supermarket sales can be analyzed in light of promotional efforts to provide knowledge of consumer buying behavior. Thus, a manufacturer or retailer could determine which items are most susceptible to promotional efforts.

### 1.1.4 Data Warehouses

Dramatic advances in data capture, processing power, data transmission, and storage capabilities are enabling organizations to integrate their various databases into *data warehouses*. Data warehousing is defined as a process of centralized data management and retrieval. Data warehousing, like data mining, is a relatively new term although the concept itself has been around for years. Data warehousing represents an ideal vision of maintaining a central repository of all organizational data.

## 1.2 What can data mining do?

Data mining is primarily used today by companies with a strong consumer focus - retail, financial, communication, and marketing organizations. It enables these companies to determine relationships among "internal" factors such as price, product positioning, or staff skills, and "external" factors such as economic indicators, competition, and customer demographics. And, it enables them to determine the impact on sales, customer satisfaction, and corporate profits. Finally, it enables them to "drill down" into summary information to view detail transactional data.

With data mining, a retailer could use point-of-sale records of customer purchases to send targeted promotions based on an individual's purchase history. By mining demographic data from comment or warranty cards, the retailer could develop products and promotions to appeal to specific customer segments. For example, Blockbuster Entertainment mines its video rental history database to recommend rentals to individual customers. American Express can suggest products to its cardholders based on analysis of their monthly expenditures.

## 1.3 How does data mining work?

While large-scale information technology has been evolving separate transaction and analytical systems, data mining provides the link between the two. Data mining software analyzes relationships and patterns in stored transaction data based on open-ended user queries. Several types of analytical software are available: statistical, machine learning, and neural networks. Generally, any of four types of relationships are sought:

- **Classes**: Stored data is used to locate data in predetermined groups. For example, a restaurant chain could mine customer purchase data to determine when customers visit and what they typically order. This information could be used to increase traffic by having daily specials.

- **Clusters**: Data items are grouped according to logical relationships or consumer preferences. For example, data can be mined to identify market segments or consumer affinities.

- **Associations**: Data can be mined to identify associations. The beer-diaper example is an example of associative mining.

- **Sequential patterns**: Data is mined to anticipate behavior patterns and trends. For example, an outdoor equipment retailer could predict the likelihood of a backpack being purchased based on a consumer's purchase of sleeping bags and hiking shoes.

Data mining process includes various internal operations as shown in Fig 1

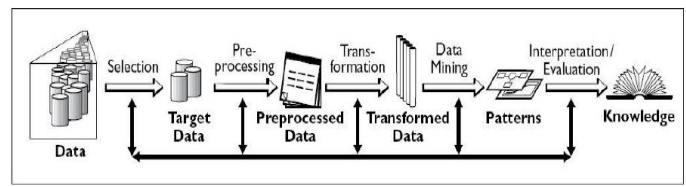

Fig 1: Data Mining Process

## 1.3.1 Data mining consists of five major elements:

- Extract, transform, and load transaction data onto the data warehouse system.

- Store and manage the data in a multidimensional database system.

- Provide data access to business analysts and information technology professionals.

- Analyze the data by application software.

- Present the data in a useful format, such as a graph or table.

## 1.3.2 Different levels of analysis are available:

- **Artificial neural networks**: Non-linear predictive models that learn through training and resemble biological neural networks in structure.

- **Genetic algorithms**: Optimization techniques that use process such as genetic combination, mutation, and natural selection in a design based on the concepts of natural evolution.

- **Decision trees**: Tree-shaped structures that represent sets of decisions. These decisions generate rules for the classification of a dataset. Specific decision tree methods include Classification and Regression Trees (CART) and Chi Square Automatic Interaction Detection (CHAID). CART and CHAID are decision tree techniques used for classification of a dataset. They provide a set of rules that you can apply to a new (unclassified) dataset to predict which records will have a given outcome. CART segments a dataset by creating 2-way splits while CHAID segments using chi square tests to create multi-way splits. CART typically requires less data preparation than CHAID.

- **Nearest neighbor method**: A technique that classifies each record in a dataset based on a combination of the classes of the $k$ record(s) most similar to it in a historical dataset (where $k$ 1). Sometimes called the $k$-nearest neighbor technique.

- **Rule induction**: The extraction of useful if-then rules from data based on statistical significance.

- **Data visualization**: The visual interpretation of complex relationships in multidimensional data. Graphics tools are used to illustrate data relationships

## 1.3.3 Operations of Web Mining:

- **Data/information extraction**: Our focus will be on extraction of structured data from Web pages, such as products and search results. Extracting such data allows one to provide services. Two main types of techniques, machine learning and automatic extraction are covered.

- **Web information integration and schema matching**: Although the Web contains a huge amount of data, each web site (or even page) represents similar information differently. How to identify or match semantically similar data is a very important problem with many practical applications. Some existing techniques and problems are examined.

- **Opinion extraction from online sources**: There are many online opinion sources, e.g., customer reviews of products, forums, blogs and chat rooms. Mining opinions (especially consumer opinions) is of great importance for marketing intelligence and product benchmarking. We will introduce a few tasks and techniques to mine such sources.

- **Knowledge synthesis**: Concept hierarchies or ontology are useful in many applications. However, generating them manually is very time consuming. A few existing methods that explores the information redundancy of the Web will be presented. The main application is to synthesize and organize the pieces of information on the Web to give the user a coherent picture of the topic domain..

- **Segmenting Web pages and detecting noise**: In many Web applications, one only wants the main content of the Web page without advertisements, navigation links, copyright notices. Automatically segmenting Web page to extract the main content of the pages is interesting problem. A number of interesting techniques have been proposed in the past few years.

## 2. Types of Web Mining

Web Mining is the extraction of interesting and potentially useful patterns and implicit information from artifacts or activity related to the World Wide Web. There are roughly three knowledge discovery domains that pertain to web mining: Web Content Mining, Web Structure Mining, and Web Usage Mining. Web content mining is the process of extracting knowledge from the content of documents or their descriptions. Web document text mining, resource discovery based on concepts indexing or agent based technology may also fall in this category. Web structure mining is the process of inferring knowledge from the World Wide Web organization and links between references and referents in the Web. Finally, web usage mining, also known as Web Log Mining, is the process of extracting interesting patterns in web access logs.

### 2.1 Web Content Mining

Web content mining is an automatic process that goes beyond keyword extraction. Since the content of a text document presents no machine-readable semantic, some approaches have suggested to restructure the document content in a representation that could be exploited by machines. The usual approach to exploit known structure in documents is to use wrappers to map documents to some data model. Techniques using lexicons for content interpretation are yet to come. There are two groups of web content mining strategies: Those that directly mine the content of documents and those that improve on the content search of other tools like search engines.

### 2.2 Web Structure Mining

World Wide Web can reveal more information than just the information contained in documents. For example, links pointing to a document indicate the popularity of the document, while links coming out of a document indicate the richness or perhaps the variety of topics covered in the document. This can be compared to bibliographical citations. When a paper is cited often, it ought to be important. The Page Rank and CLEVER methods take advantage of this information conveyed by the links to find pertinent web pages.

### 2.3 Web Usage Mining

Web servers record and accumulate data about user interactions whenever requests for resources are received. Analyzing the web access logs of deferent web sites can help understand the user behavior and the web structure, thereby improving the design of this colossal collection of resources. There are two main tendencies in Web Usage Mining driven by the applications of the discoveries: General Access Pattern Tracking and Customized Usage Tracking. The general access pattern tracking analyzes the web logs to understand access patterns and trends. These analyses can shed light on better structure and grouping of resource providers. Many web analysis tools existed but they are limited and usually unsatisfactory. We have designed a web log data mining tool, Weblog Miner, and proposed techniques for using data mining and Online Analytical Processing (OLAP) on treated and transformed web access files. Some scripts custom-tailored for some sites may store additional information. However, for an effective web usage mining, an important

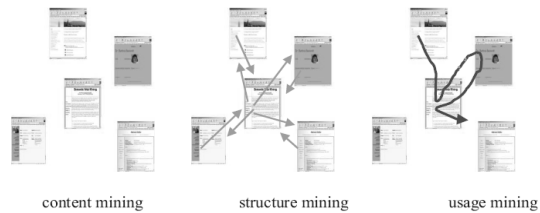

content mining    structure mining    usage mining

Fig 2 : Types of Web Mining

## 3. What is Semantic Web Mining?

The effort behind the Semantic Web is to add semantic annotation to Web documents in order to access knowledge instead of unstructured material, allowing knowledge to be managed in an automatic way. Web Mining can help to learn definitions of structures for knowledge organization (e. g., ontologies) and to provide the population of such knowledge structures.

All approaches discussed here are semi-automatic. They assist the knowledge engineer in extracting the semantics, but cannot completely replace her. In order to obtain high-quality results, one cannot replace the human in the loop, as there is always a lot of tacit knowledge involved in the modeling process. A computer will never be able to fully consider background knowledge, experience, or social conventions.

### 3.1 Ontology Learning:

Extracting ontology from the Web is a challenging task. One way is to engineer the ontology by hand, but this is quite an expensive way. The expression Towards Semantic Web Mining *Ontology Learning* was coined for the semi-automatic extraction of semantics from the Web in order to create ontology the techniques produce intermediate results which must finally be integrated in one machine-understandable format, ex: ontology.

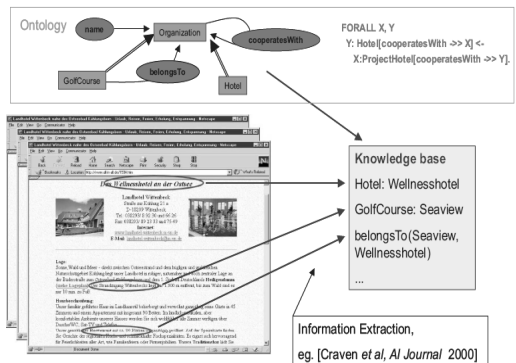

Fig 3: Ontology Mapping and Merging

### 3.2 Mapping and Merging Ontologies

With the growing usage of ontologies, the problem of overlapping knowledge in a common domain occurs more often and becomes critical. Domain-specific ontologies are modeled by multiple authors in multiple settings. These ontologies lay the foundation for building new domain-specific ontologies in similar domains by assembling and extending multiple ontologies from repositories. The process of *ontology merging* takes as input two (or more) source ontologies and returns a merged ontology based on the given source ontologies. Another method is FCA-Merge which merges ontologies following a bottom-up approach, offering a global structural description of the process.

### 4. Architecture of Semantic Web Miner

In this semantic web miner architecture includes various components to perform syntactic as well as semantic analysis on the web resources for the given input string. At the same time each web server must contains one special module Meta Information Analyzer.

### 5. Features of Semantic Web Miner

Semantic web Miner Marjory considers the concept of time complexity of searching. In general search engines will need to spend more time in searching the web results by simple matching of given input string with the contents of each and every web resource

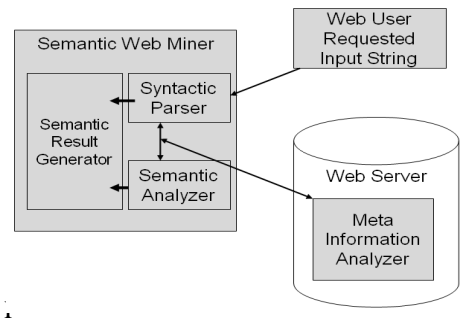

Fig 4: Semantic Web Miner

## 6. What is Neural Network?

An Artificial Neural Network (ANN) is an information processing paradigm that is inspired by the way biological nervous systems, such as the brain, process information. The key element of this paradigm is the novel structure of the information processing system. It is composed of a large number of highly interconnected processing elements (neurons) working in unison to solve specific problems. ANNs, like people, learn by example.

### 6.1 Why use neural networks?

Neural networks, with their remarkable ability to derive meaning from complicated or imprecise data, can be used to extract patterns and detect trends that are too complex to be noticed by either humans or other computer techniques.

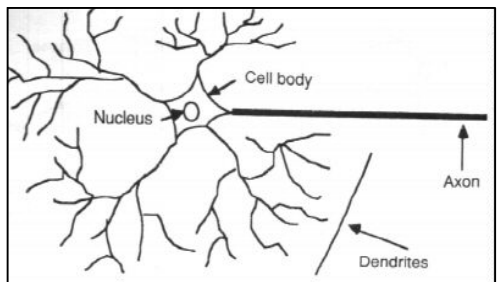

Fig 5: Components of a neuron

Other advantages include:

1. Adaptive learning: An ability to learn how to do tasks based on the data given for training or initial experience.

2. Self-Organization: An ANN can create its own organization or representation of the information it receives during learning time.

Real Time Operation: ANN computations may be carried out in parallel, and special

hardware devices are being designed and

3. manufactured which take advantage of this capability.

4. Fault Tolerance via Redundant Information Coding: Partial destruction of a network leads to the corresponding degradation of performance. However, some network capabilities may be retained even with major network damage.

### 6.2 How the Human Brain Learns?

In the human brain, a typical neuron collects signals from others through a host of fine structures called *dendrites*. The neuron sends out spikes of electrical activity through a long, thin stand known as an *axon*, which splits into thousands of branches. At the end of each branch, a structure called a *synapse* converts the activity from the axon into electrical effects that inhibit or excite activity from the axon into electrical effects that inhibit or excite activity in the connected neurons. When a neuron receives excitatory input that is sufficiently large  compared with its inhibitory input, it sends a spike of electrical activity down its axon.

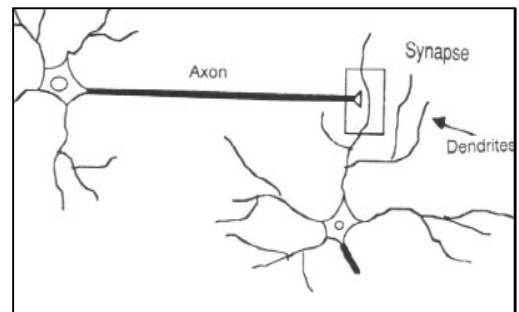

Fig 6: Synapse

Learning occurs by changing the effectiveness of the synapses so that the influence of one neuron on another changes.

### 6.3 From Human Neurons to Artificial Neurons

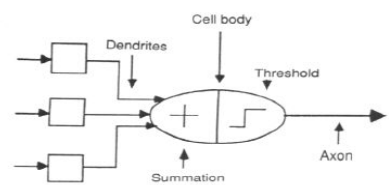

Fig 8: Neuron Structure

We conduct these neural networks by first trying to deduce the essential features of neurons and their interconnections.

## 7. Components of Artificial Neural Analyzer

### 7.1 A simple neuron

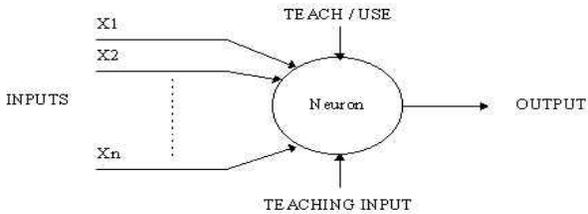

Fig 8: Simple Neuron

An artificial neuron is a device with many inputs and one output. The neuron has two modes of operation; the training mode and the using mode. In the training mode, the neuron can be trained to fire (or not), for particular input patterns.

### 7.2 A more complicated neuron

The previous neuron doesn't do anything that conventional computers don't do already. A more sophisticated neuron is the McCulloch and Pitts model (MCP). The difference from the previous model is that the inputs are 'weighted'; the effect that each input has at decision making is dependent on the weight of the particular input. The weight of an input is a number which when multiplied with the input gives the weighted input. In any other case the neuron does not fire. In mathematical terms, the neuron fires if and only if;

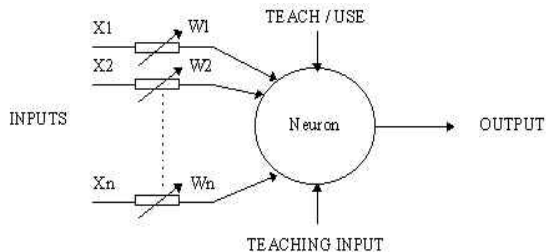

Fig 9: Complicated Neuron

$$X1W1 + X2W2 + X3W3 + ... > T$$

The addition of input weights and of the threshold makes this neuron a very flexible and powerful one. The MCP neuron has the ability to adapt to a particular situation by changing its weights and/or threshold. Various algorithms exist that cause the neuron to 'adapt'; the most used ones are the Delta rule and the back error propagation. The former is used in feed-forward networks and the latter in feedback networks.

### 7.3 Feed-forward networks

Feed-forward ANNs allow signals to travel one way only; from input to output. There is no feedback (loops) i.e. the output of any layer does not affect that same layer. Feed-forward ANNs tend to be straight forward networks that associate inputs with outputs. They are extensively used in pattern recognition. This type of organization is also referred to as bottom-up or top-down.

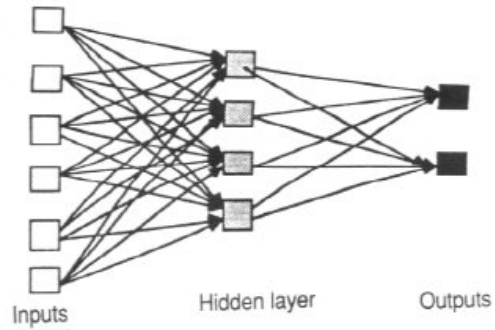

Fig 10: General Neural Network

### 7.3.1 Network layers

The commonest type of artificial neural network consists of three groups, or layers, of units: a layer of "**input**" units is connected to a layer of "**hidden**" units, which is connected to a layer of "**output**" units.

- The activity of the input units represents the raw information that is fed into the network.

- The activity of each hidden unit is determined by the activities of the input units and the

weights on the connections between the input and the hidden units.

• The behavior of the output units depends on the activity of the hidden units and the weights between the hidden and output units.

This simple type of network is interesting because the hidden units are free to construct their own representations of the input. The weights between the input and hidden units determine when each hidden unit is active, and so by modifying these weights, a hidden unit can choose what it represents.

We also distinguish single-layer and multi-layer architectures. The single-layer organization, in which all units are connected to one another, constitutes the most general case and is of more potential computational power than hierarchically structured multi-layer organizations.

## 7.4 Feedback networks

Feedback networks can have signals traveling in both directions by introducing loops in the network. Feedback networks are very powerful and can get extremely complicated. Feedback networks are dynamic; their 'state' is changing continuously until they reach an equilibrium point.

## 7.5 Back-propagation Algorithm - a mathematical approach

Units are connected to one another. Connections correspond to the edges of the underlying directed graph. There is a real number associated with each connection, which is called the weight of the connection. We denote by Wij the weight of the connection from unit ui to unit uj. It is then convenient to represent the pattern of connectivity in the network by a weight matrix W whose elements are the weights Wij. Two types of connection are usually distinguished: excitatory and inhibitory. A positive weight represents an excitatory connection whereas a negative weight represents an inhibitory connection

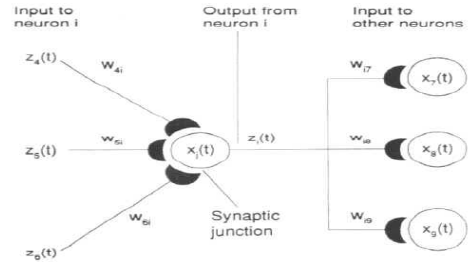

Fig 11: Back Propagation Neural Network

A unit in the output layer determines its activity by following a two step procedure.

• First, it computes the total weighted input xj, using the formula:

$$X_j = \sum_i y_i W_{ij}$$

Where yi is the activity level of the jth unit in the previous layer and Wij is the weight of the connection between the ith and the jth unit.

• Next, the unit calculates the activity yj using some function of the total weighted input. Typically we use the sigmoid function:

$$y_j = \frac{1}{1 + e^{-x_j}}$$

• Once the activities of all output units have been determined, the network computes the error E, which is defined by the expression:

$$E = \frac{1}{2} \sum_i \left(y_i - d_i\right)^2$$

Where yj is the activity level of the jth unit in the top layer and dj is the desired output of the jth unit.

### 7.5.1 The back-propagation algorithm consists of four steps:

1. Compute how fast the error changes as the activity of an output unit is changed. This error derivative (EA) is the difference between the actual and the desired activity.

$$EA_j = \frac{\partial E}{\partial y_j} = y_j - d_j$$

2. Compute how fast the error changes as the total input received by an output unit is changed. This quantity (EI) is the answer from step 1 multiplied by the rate at which the output of a unit changes as its total input is changed.

$$EI_j = \frac{\partial E}{\partial x_j} = \frac{\partial E}{\partial y_j} \times \frac{dy_j}{dx_j} = EA_j y_j \left(1 - y_j\right)$$

3. Compute how fast the error changes as a weight on the connection into an output unit is changed. This quantity (EW) is the answer from step 2 multiplied by the activity level of the unit from which the connection emanates.

$$EW_{ij} = \frac{\partial E}{\partial W_{ij}} = \frac{\partial E}{\partial x_j} \times \frac{\partial x_j}{\partial W_{ij}} = EI_j y_i$$

4. Compute how fast the error changes as the activity of a unit in the previous layer is changed. It is the answer in step 2 multiplied by the weight on the connection to that output unit.

$$EA_i = \frac{\partial E}{\partial y_i} = \sum_j \frac{\partial E}{\partial x_j} \times \frac{\partial x_j}{\partial y_i} = \sum_j EI_j W_{ij}$$

By using steps 2 and 4, we can convert the EAs of one layer of units into EAs for the previous layer. This procedure can be repeated to get the EAs for as many previous layers as desired. Once we know the EA of a unit, we can use steps 2 and 3 to compute the EWs on its incoming connections.

## 8. Features of Artificial Neural Analyzer

Depending on the nature of the application and the strength of the internal data patterns you can generally expect a network to train quite well. This applies to problems where the relationships may be quite dynamic or non-linear. ANNs provide an analytical alternative to conventional techniques which are often limited by strict assumptions of normality, linearity, variable independence etc. Some of the major advantages of Artificial Neural Analyzer are:

- There is no need to assume an underlying data distribution such as usually is done in statistical modeling.
- Neural networks are applicable to multivariate non-linear problems.
- The transformations of the variables are automated in the computational process

## 9. Components of Probabilistic Semantic Web Mining

This paper proposes the idea of combining the efficient methods those can make efficient for the web mining. For obtaining this effect, search engine should mainly contain three components:

- Semantic Web Miner,
- Artificial Neural Analyzer, &
- Result Formatter.

### 9.1 Semantic Web Miner

As we have already seen about working of general semantic web mining module. This semantic web miner will include: Syntactic Parser, Semantic Analyzer and Semantic Result Generator. This miner performs first level of web results analysis.

### 9.1.1 Syntactic Parser: it is simple parser that reads the given input string, and converts to corresponding syntax tree. This syntax tree can be used for calculating the syntactic matched results. Large collection of web results will obtain with this simple syntactic parser. Most of these results may not useful, so it will be sending for semantic analysis.

### 9.1.2 Semantic Analyzer: this analyzer will verify the semantic details of each and every result obtained in the syntactic parser. This analyzer will verify the semantics by contacting the web server Meta information details of the web resources.

### 9.1.3 Semantic Result Generator: this component is useful for finalizing the semantically

matched web result from the list of syntactically matched web results. Even though these results may contains some of the un wanted results in miner levels.

## 9.2 Artificial Neural Analyzer (ANA):

This will be the second level of web analysis. This analyzer contains various components required for producing Artificial Neural Network effect. Along with that artificial neural network, it also included with Probabilities attacher. This probabilities attacher will add probability labels for each and every semantic result that are obtained from the ANA..

## 9.3 Result Formatter:

This formatter contains various methods of representing the output result. Since probabilities play major role in these analyses, it must sort out the web results according to its probabilities order. Those sorting algorithms must be included in this formatter at the same time it must also include probability cut off analysis in order to restrict the low probability web results.

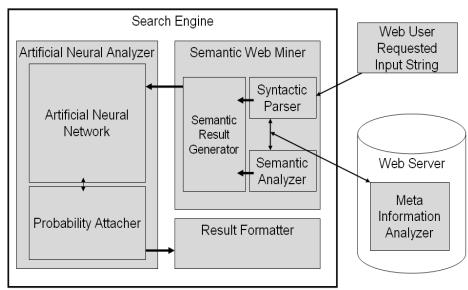

Fig 12: Probabilistic Semantic Web Miner with Artificial Neural Analysis Architecture

When if a web user requested to search a string "semantic web mining", then syntactic analyzer will construct syntax tree with three leaf nodes labeled "semantic", "web", and "mining". Then search for the matching on the web server.

## 10. Conclusion

The idea proposed in this paper can give powerful architecture for the search engines in web mining. Semantic web miner used in this architecture will help in reducing the time required for searching results. Probabilities attached to the semantic results can be used to reduce number or count of un wanted result. With this combined techniques will give more accurate results of web resources with in less time. This paper idea that attached to search engines will produce only relevant information in order to reduce burden of un wanted results for the web user request to same their valuable search time.

## 11. References

[1] B. Berendt. Using site semantics to analyze, visualize and support navigation. Data Mining and Knowledge Discovery, 6:37–59, 2002.

]2] B. Berendt and M. Spiliopoulou. Analysing navigation behaviour in web sites integrating multiple information systems. The VLDB Journal, 9(1):56–75, 2000.

[3] S. Chakrabarti. Data mining for hypertext: Atutorial survey. SIGKDD Explorations, 1:1–11, 2000.

[4] S. Chakrabarti, B. Dom, D. Gibson, J. Kleinberg, P. Raghavan, and S. Rajagopalan.

[5] S. Chakrabarti, M. van den Berg, and B. Dom.

[6] Hans Chalupsky. Ontomorph: Atranslation system for symbolic knowledge.

[7] G. Chang, M.J. Healey, J.A.M. McHugh, and J.T.L.Wang. Mining the World Wide Web. An Information Search Approach.

[8] E.H. Chi, P. Pirolli, and J. Pitkow. The scent of a site: a system for analyzing and predicting information scent, usage, and usability of a web site.

## AUTHORS

**1. T. KRISHNA KISHORE**, working as an Asst. Professor, Dept. of CSE, in ST. ANN'S COLLEGE OF ENGINEERING & TECHNOLOGY,CHIRALA-523187,PRAKASAM (DT.), ANDHRA PRADESH, INDIA.

His Interested Areas are COMPUTER NETWORKS, INFORMATION SECURITY, THEORY OF COMPUTATION, COMPILER DESIGN, EMBEDDED SYSTEMS…..

Mobile: + 91 – 9885406051

E-mail: thota_btech@yahoo.com

**2. T. SASI VARDHAN**, working as an Asst. Professor, Dept. of IT, in ST. ANN'S ENGINEERING COLLEGE, CHIRALA-523187,PRAKASAM (DT.), ANDHRA PRADESH, INDIA.

His Interested Areas are COMPUTER NETWORKS, INFORMATION SECURITY, DATA MINING & WAREHOUSING, and SOFTWARE ENGINEERING…..

Mobile: +91 – 9959859922

E-m ail: sasi_thota36@yahoo.co.in

**3. N. LAKSHMI NARAYANA**, working as an Asst. Professor, Dept. of CSE, in ST. ANN'S COLLEGE OF ENGINEERING & TECHNOLOGY,CHIRALA-523187,PRAKASAM (DT.), ANDHRA PRADESH, INDIA.

His Interested Areas are COMPUTER NETWORKS, WEB DESIGNING, COMPILER DESIGN, TESTING METHODOLOGIES…..

Mobile: +91 – 9030136208

E-mail: thisisnarayana@gmail.com